\documentclass[11pt,a4paper]{article}
\usepackage[hyperref]{emnlp2020}
\usepackage{times}
\usepackage{latexsym}

\usepackage{amsmath}
\usepackage{amsfonts}
\usepackage{multirow,multicol}
\usepackage{pifont}
\newcommand{\cmark}{\ding{51}}%
\newcommand{\xmark}{\ding{55}}%
\usepackage{graphicx}

\usepackage{microtype}

\aclfinalcopy 

\title{Efficient Context and Schema Fusion Networks for Multi-Domain Dialogue State Tracking}

\author{Su Zhu, Jieyu Li, Lu Chen\footnotemark[1] \and Kai Yu\thanks{\ \ The corresponding authors are Lu Chen and Kai Yu.}\\
  State Key Laboratory of Media Convergence Production Technology and Systems\\
  MoE Key Lab of Artificial Intelligence, AI Institute, Shanghai Jiao Tong University\\
  SpeechLab, Department of Computer Science and Engineering\\
  Shanghai Jiao Tong University, Shanghai, China\\
  {\tt \{paul2204,oracion,chenlusz,kai.yu\}@sjtu.edu.cn} \\}

\date{}

\begin{document}
\maketitle

\begin{abstract}
Dialogue state tracking (DST) aims at estimating the current dialogue state given all the preceding conversation. For multi-domain DST, the data sparsity problem is a major obstacle due to increased numbers of state candidates and dialogue lengths. To encode the dialogue context efficiently, we utilize the previous dialogue state (predicted) and the current dialogue utterance as the input for DST. To consider relations among different domain-slots, the schema graph involving prior knowledge is exploited. In this paper, a novel context and schema fusion network is proposed to encode the dialogue context and schema graph by using internal and external attention mechanisms. Experiment results show that our approach can outperform strong baselines, and the previous state-of-the-art method (SOM-DST) can also be improved by our proposed schema graph.



\end{abstract}

\section{Introduction}
\label{sec:intro}

\begin{figure}
    \centering
    \includegraphics[width=0.99\linewidth]{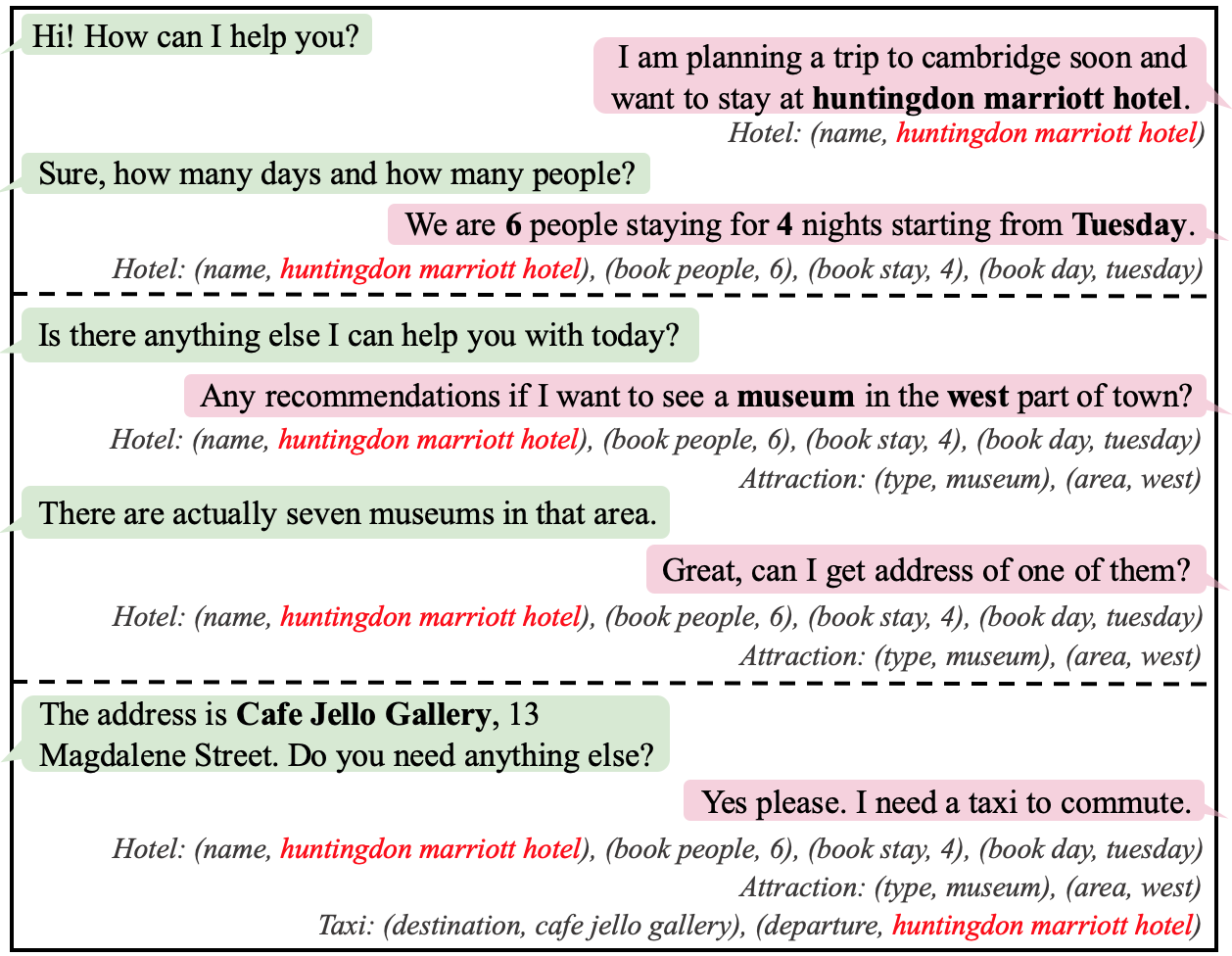}
    \caption{An example of multi-domain dialogues. Utterances at the left side are from the system agent, and utterances at the right side are from a user. The dialogue state of each domain is represented as a set of (\emph{slot}, \emph{value}) pairs.} 
    \label{fig:dial_example}
\end{figure}

Dialogue state tracking (DST) is a key component in task-oriented dialogue systems which cover certain narrow domains (e.g., \emph{booking hotel} and \emph{travel planning}). As a kind of context-aware language understanding task, DST aims to extract user goals or intents hidden in human-machine conversation and represent them as a compact dialogue state, i.e., a set of slots and their corresponding values. For example, as illustrated in Fig. \ref{fig:dial_example}, (\emph{slot}, \emph{value}) pairs like (\emph{name}, \emph{huntingdon marriott hotel}) are extracted from the dialogue. It is essential to build an accurate DST for dialogue management \cite{young2013pomdp}, where dialogue state determines the next machine action and response. 

Recently, motivated by the tremendous growth of commercial dialogue systems like Apple Siri, Microsoft Cortana, Amazon Alexa, or Google Assistant, multi-domain DST becomes crucial to help users across different domains~\cite{budzianowski-etal-2018-multiwoz,eric2019multiwoz}. As shown in Fig. \ref{fig:dial_example}, the dialogue covers three domains (i.e., \texttt{Hotel}, \texttt{Attraction} and \texttt{Taxi}). The goal of multi-domain DST is to predict the value (including \texttt{NONE}) for each \emph{domain-slot} pair based on all the preceding dialogue utterances. However, due to increasing numbers of dialogue turns and domain-slot pairs, the data sparsity problem becomes the main issue in this field.

To tackle the above problem, we emphasize that DST models should support open-vocabulary based value decoding, encode context efficiently and incorporate domain-slot relations:
\begin{enumerate}
    \item Open-vocabulary DST is essential for real-world applications~\cite{wu2019transferable,gao-etal-2019-dialog,ren-etal-2019-scalable}, since value sets for some slots can be very huge and variable (e.g., \emph{song names}). 
    \item To encode the dialogue context efficiently, we attempt to get context representation from the previous (predicted) dialogue state and the current turn dialogue utterance, while not concatenating all the preceding dialogue utterances.
    \item To consider relations among domains and slots, we introduce the schema graph which contains \emph{domain}, \emph{slot}, \emph{domain-slot} nodes and their relationships. It is a kind of prior knowledge and may help alleviate the data imbalance problem.
\end{enumerate}

To this end, we propose a multi-domain dialogue state tracker with context and schema fusion networks (CSFN-DST). The fusion network is exploited to jointly encode the previous dialogue state, the current turn dialogue and the schema graph by internal and external attention mechanisms. After multiple layers of attention networks, the final representation of each \emph{domain-slot} node is utilized to predict the corresponding value, involving context and schema information. For the value prediction, a slot gate classifier is applied to decide whether a domain-slot is mentioned in the conversation, and then an RNN-based value decoder is exploited to generate the corresponding value.

Our proposed CSFN-DST is evaluated on MultiWOZ 2.0 and MultiWOZ 2.1 benchmarks. Ablation study on each component further reveals that both context and schema are essential. Contributions in this work are summarized as:
\begin{itemize}
    \item To alleviate the data sparsity problem and enhance the context encoding, we propose exploiting domain-slot relations within the schema graph for open-vocabulary DST.
    \item To fully encode the schema graph and dialogue context, fusion networks are introduced with graph-based, internal and external attention mechanisms.
    \item Experimental results show that our approach surpasses strong baselines, and the previous state-of-the-art method (SOM-DST) can also be improved by our proposed schema graph.
\end{itemize}


\section{Related Work}
\label{sec:rel}

Traditional DST models rely on semantics extracted by natural language understanding to predict the current dialogue states \cite{young2013pomdp,williams-EtAl:2013:SIGDIAL,henderson-slt14.1,sun2014sjtu,sun-slt14,yu2015constrained}, or jointly learn language understanding in an end-to-end way \cite{henderson-slt14.2,henderson-thomson-young:2014:W14-43}. These methods heavily rely on hand-crafted features and complex domain-specific lexicons for delexicalization, which are difficult to extend to new domains. Recently, most works about DST focus on encoding dialogue context with deep neural networks (such as CNN, RNN, LSTM-RNN, etc.) and predicting a value for each possible slot \cite{mrkvsic2017neural,xu2018acl,zhong2018global,ren2018towards}. 

\noindent\textbf{Multi-domain DST}\quad Most traditional state tracking approaches focus on a single domain, which extract value for each slot in the domain \cite{williams-EtAl:2013:SIGDIAL,henderson-thomson-williams:2014:W14-43}. They can be directly adapted to multi/mixed-domain conversations by replacing slots in a single domain with \emph{domain-slot} pairs (i.e. domain-specific slots)~\cite{ramadan2018large,gao-etal-2019-dialog,wu2019transferable,zhang2019find,kim2019efficient}. Despite its simplicity, this approach for multi-domain DST extracts value for each domain-slot independently, which may fail to capture features from slot co-occurrences. For example, hotels with higher \emph{stars} are usually more expensive (\emph{price\_range}).



\noindent\textbf{Predefined ontology-based DST}\quad Most of the previous works assume that a predefined ontology is provided in advance, i.e., all slots and their values of each domain are known and fixed~\cite{williams2012belief,henderson-thomson-williams:2014:W14-43}. Predefined ontology-based DST can be simplified into a value classification task for each slot~\cite{henderson-thomson-young:2014:W14-43,mrkvsic2017neural,zhong2018global,ren2018towards,ramadan2018large,lee2019sumbt}. It has the advantage of access to the known candidate set of each slot, but these approaches may not be applicable in the real scenario. Since a full ontology is hard to obtain in advance \cite{xu2018acl}, and the number of possible slot values could be substantial and variable (e.g., \emph{song names}), even if a full ontology exists~\cite{wu2019transferable}.

\noindent\textbf{Open-vocabulary DST}\quad Without a predefined ontology, some works choose to directly generate or extract values for each slot from the dialogue context, by using the encoder-decoder architecture \cite{wu2019transferable} or the pointer network \cite{gao-etal-2019-dialog,ren-etal-2019-scalable,Le2020Non-Autoregressive}. They can improve the scalability and robustness to unseen slot values, while most of them are not efficient in context encoding since they encode all the previous utterances at each dialogue turn. Notably, a multi-domain dialogue could involve quite a long history, e.g., MultiWOZ dataset~\cite{budzianowski-etal-2018-multiwoz} contains about 13 turns per dialogue on average.

\noindent\textbf{Graph Neural Network}\quad Graph Neural Network (GNN) approaches~\cite{scarselli2009graph,velickovic2018graph} aggregate information from graph structure and encode node features, which can learn to reason and introduce structure information. Many GNN variants are proposed and also applied in various NLP tasks, such as text classification \cite{yao2019graph}, machine translation \cite{marcheggiani2018exploiting}, dialogue policy optimization \cite{chen2018structured,chen2019agentgraph} etc. We introduce graph-based multi-head attention and fusion networks for encoding the schema graph.

\section{Problem Formulation}

In a multi-domain dialogue state tracking problem, we assume that there are $M$ domains (e.g. \emph{taxi}, \emph{hotel}) involved, $\mathcal{D}=\{d_1,d_2,\cdots,d_M\}$. Slots included in each domain $d\in\mathcal{D}$ are denoted as a set $\mathcal{S}^d=\{s_1^d,s_2^d,\cdots,s_{|\mathcal{S}^d|}^d\}$.\footnote{For open-vocabulary DST, possible values for each slot $s\in\mathcal{S}^d$ are not known in advance.} Thus, there are $J$ possible \emph{domain}-\emph{slot} pairs totally, $\mathcal{O}=\{O_1,O_2,\cdots,O_J\}$, where $J=\sum_{m=1}^M |\mathcal{S}^{d_m}|$. Since different domains may contain a same slot, we denote all distinct $N$ slots as $\mathcal{S}=\{s_1,s_2,\cdots,s_N\}$, where $N\leq J$.

A dialogue can be formally represented as $\{(A_1,U_1,B_1),(A_2,U_2,B_2),\cdots,(A_T,U_T,B_T)\}$, where $A_t$ is what the agent says at the $t$-th turn, $U_t$ is the user utterance at $t$ turn, and $B_t$ denotes the corresponding dialogue state. $A_t$ and $U_t$ are word sequences, while $B_t$ is a set of \emph{domain}-\emph{slot}-\emph{value} triplets, e.g., (\emph{hotel}, \emph{price\_range}, \emph{expensive}). Value $v_{tj}$ is a word sequence for $j$-th \emph{domain}-\emph{slot} pair at the $t$-th turn. The goal of DST is to correctly predict the value for each \emph{domain}-\emph{slot} pair, given the dialogue history.

Most of the previous works choose to concatenate all words in the dialogue history, $[A_1,U_1,A_2,U_2,\cdots,A_t,U_t]$, as the input. However, this may lead to increased computation time. In this work, we propose to utilize only the current dialogue turn $A_t$, $U_t$ and the previous dialogue state $B_{t-1}$ to predict the new state $B_t$. During the training, we use the ground truth of $B_{t-1}$, while the previous predicted dialogue state would be used in the inference stage.

\begin{figure}
    \centering
    \includegraphics[width=0.9\linewidth]{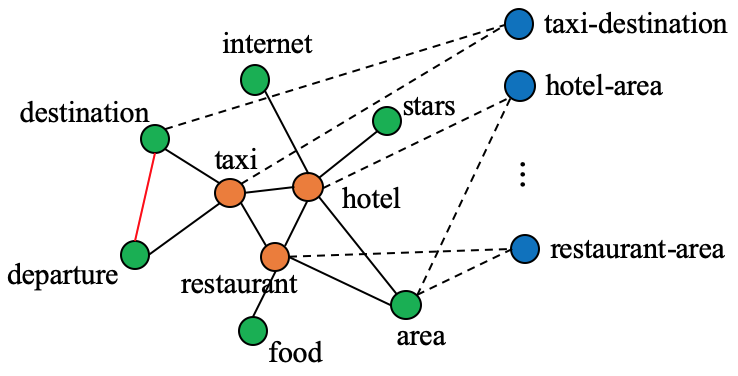}
    \caption{An example of schema graph. Domain nodes are in orange, slot nodes are in green and domain-slot nodes are in blue.}
    \label{fig:schema_graph}
\end{figure}

\noindent\textbf{Schema Graph}\quad To consider relations between different \emph{domain}-\emph{slot} pairs and exploit them as an additional input to guide the context encoding, we formulate them as a schema graph $G=(V,E)$ with node set $V$ and edge set $E$. Fig. \ref{fig:schema_graph} shows an example of schema graph. In the graph, there are three kinds of nodes to denote all domains $\mathcal{D}$, slots $\mathcal{S}$, and domain-slot pairs $\mathcal{O}$, i.e., $V=\mathcal{D}\cup\mathcal{S}\cup\mathcal{O}$. Four types of undirected edges between different nodes are exploited to encode prior knowledge:
\begin{enumerate}
    \item $(d, d^{\prime})$: Any two domain nodes, $d\in\mathcal{D}$ and $d^{\prime}\in\mathcal{D}$, are linked to each other.
    \item $(s, d)$: We add an edge between slot $s\in\mathcal{S}$ and domain $d\in\mathcal{D}$ nodes if $s\in\mathcal{S}^{d}$, .
    \item $(d, o)$ and $(s, o)$: If a domain-slot pair $o\in\mathcal{O}$ is composed of the domain $d\in\mathcal{D}$ and slot $s\in\mathcal{S}$, there are two edges from $d$ and $s$ to this domain-slot node respectively.
    \item $(s, s^{\prime})$: If the candidate values of two different slots ($s\in\mathcal{S}$ and $s^{\prime}\in\mathcal{S}$) would overlap, there is also an edge between them, e.g., \emph{destination} and \emph{departure}, \emph{leave\_at} and \emph{arrive\_by}.
\end{enumerate}

\section{Context and Schema Fusion Networks for Multi-domain DST}
\label{sec:method}

\begin{figure*}
    \centering
    \includegraphics[width=0.9\linewidth]{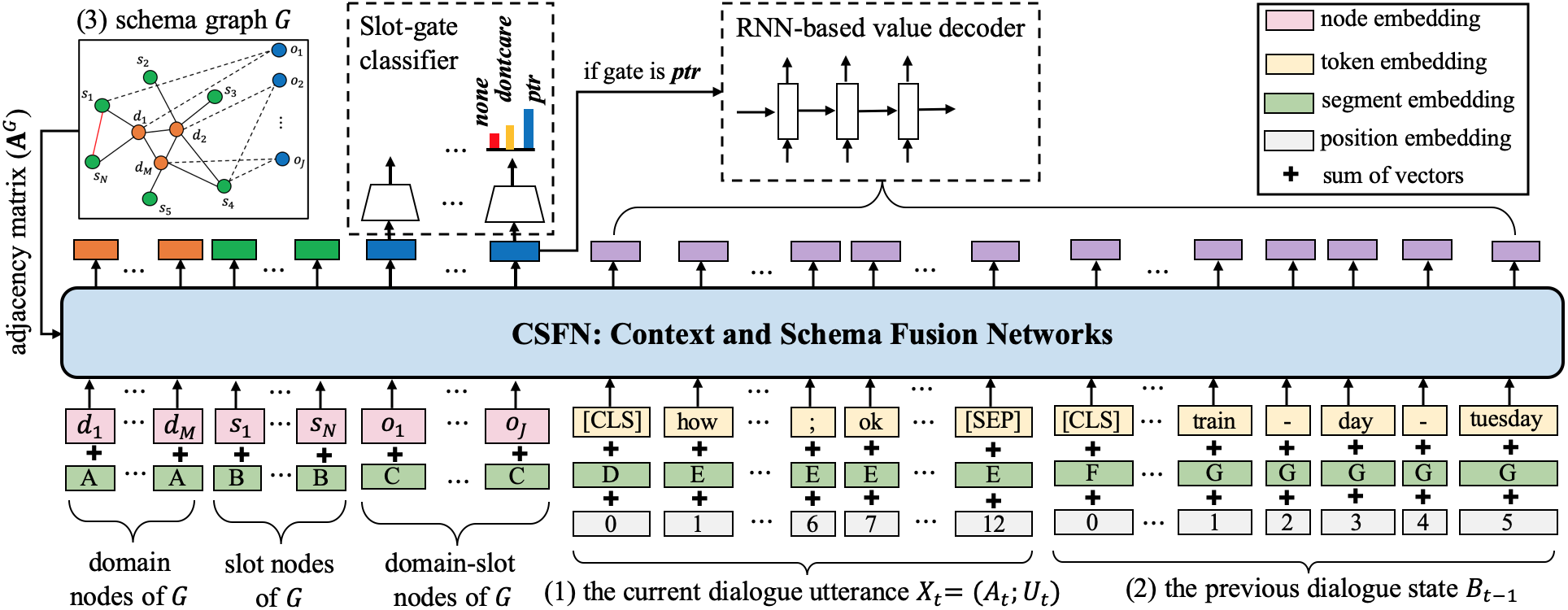}
    \caption{The overview of the proposed CSFN-DST. It takes the current dialogue utterance, the previous dialogue state and the schema graph as the input and predicts the current dialogue state. It consists of an embedding layer, context and schema fusion networks, a slot-gate classifier and an RNN-based value decoder.}
    \label{fig:model_arch}
\end{figure*}

In this section, we will introduce our approach for multi-domain DST, which jointly encodes the current dialogue turn ($A_t$ and $U_t$), the previous dialogue state $B_{t-1}$ and the schema graph $G$ by fusion networks. After that, we can obtain context-aware and schema-aware node embeddings for all $J$ domain-slot pairs. Finally, a slot-gate classifier and RNN-based value decoder are exploited to extract the value for each domain-slot pair.

The architecture of CSFN-DST is illustrated in Fig. \ref{fig:model_arch}, which consists of input embeddings, context schema fusion network and state prediction modules.


\subsection{Input Embeddings}

Besides token and position embeddings for encoding literal information, segment embeddings are also exploited to discriminate different types of input tokens.

\noindent\textbf{(1) Dialogue Utterance}\quad We denote the representation of the dialogue utterances at $t$-th turn as a joint sequence, $X_t=\texttt{[CLS]} \oplus A_{t} \oplus ; \oplus U_{t} \oplus\texttt{[SEP]}$, where $\texttt{[CLS]}$ and $\texttt{[SEP]}$ are auxiliary tokens for separation, $\oplus$ is the operation of sequence concatenation. As $\texttt{[CLS]}$ is designed to capture the sequence embedding, it has a different segment type with the other tokens. The input embeddings of $X_t$ are the sum of the token embeddings, the segmentation embeddings and the position embeddings~\cite{vaswani2017attention}, as shown in Fig. \ref{fig:model_arch}.

\noindent\textbf{(2) Previous Dialogue State}\quad As mentioned before, a dialogue state is a set of \emph{domain}-\emph{slot}-\emph{value} triplets with a mentioned value (not \texttt{NONE}). Therefore, we denote the previous dialogue state as $B_{t-1}=\texttt{[CLS]} \oplus R_{t-1}^1 \oplus \cdots \oplus R_{t-1}^{K}$, where $K$ is the number of triplets in $B_{t-1}$. Each triplet $d\texttt{-}s\texttt{-}v$ is denoted as a sub-sequence, i.e., $R=d\oplus\texttt{-}\oplus s \oplus\texttt{-}\oplus v$. The domain and slot names are tokenized, e.g., \emph{price\_range} is replaced with ``price range''. The value is also represented as a token sequence. For the special value \texttt{DONTCARE} which means users do not care the value, it would be replaced with ``dont care''. The input embeddings of $B_{t-1}$ are the sum of the token, segmentation and position embeddings. Positions are re-enumerated for different triplets. 

\noindent\textbf{(3) Schema Graph}\quad As mentioned before, the schema graph $G$ is comprised of $M$ domain nodes, $N$ slot nodes and $J$ domain-slot nodes. These nodes are arranged as $G=d_1 \oplus \cdots \oplus d_M \oplus s_1 \oplus \cdots \oplus s_N \oplus o_1 \oplus \cdots \oplus o_J$. Each node embedding is initialized by averaging embeddings of tokens in the corresponding domain/slot/domain-slot. Positions embeddings are omitted in the graph. The edges of the graph are represented as an adjacency matrix $\textbf{A}^G$ whose items are either one or zero, which would be used in the fusion network. To emphasize edges between different types of nodes can be different in the computation, we exploit node types to get segment embeddings.

\subsection{Context and Schema Fusion Network}

At this point, we have input representations $\mathrm{H}_0^{G}\in \mathbb{R}^{|G| \times d_{m}}, \mathrm{H}_0^{X_t}\in \mathbb{R}^{|X_t| \times d_{m}}, \mathrm{H}_0^{B_{t-1}}\in \mathbb{R}^{|B_{t-1}| \times d_{m}}$, where $|.|$ gets the token or node number. The context and schema fusion network (CSFN) is utilized to compute hidden states for tokens or nodes in $X_t$, $B_{t-1}$ and $G$ layer by layer. We then apply a stack of $L$ context- and schema-aware self-attention layers to get final hidden states, $\mathrm{H}_L^{G}, \mathrm{H}_L^{X_t}, \mathrm{H}_L^{B_{t-1}}$. The $i$-th layer ($0\leq i < L$) can be formulated as:
{
\begin{equation*}
    \mathrm{H}_{i+1}^{G}, \mathrm{H}_{i+1}^{X_t}, \mathrm{H}_{i+1}^{B_{t-1}} = \text{\small CSFNLayer}_i(\mathrm{H}_{i}^{G}, \mathrm{H}_{i}^{X_t}, \mathrm{H}_{i}^{B_{t-1}})
\end{equation*}}


\subsubsection{Multi-head Attention}

Before describing the fusion network, we first introduce the multi-head attention \cite{vaswani2017attention} which is a basic module. The multi-head attention can be described as mapping a query and a set of key-value pairs to an output, where the query, keys, values, and output are all vectors. The output is computed as a weighted sum of the values, where the weight assigned to each value is computed by a compatibility function of the query with the corresponding key. 

Consider a source sequence of vectors $Y=\{\boldsymbol{y}_i\}_{i=1}^{|Y|}$ where $\boldsymbol{y}_i \in \mathbb{R}^{1\times d_{\text{model}}}$ and $Y \in \mathbb{R}^{|Y| \times d_{\text{model}}}$, and a target sequence of vectors $Z=\{\boldsymbol{z}_i\}_{i=1}^{|Z|}$ where $\boldsymbol{z}_i \in \mathbb{R}^{1\times d_{\text{model}}}$ and $Z \in \mathbb{R}^{|Z| \times d_{\text{model}}}$. For each vector $\boldsymbol{y}_i$, we can compute an attention vector $\boldsymbol{c}_i$ over $Z$ by using $H$ heads as follows:
{\small
\begin{align*}
 e_{ij}^{(h)} &= \frac{(\boldsymbol{y}_i W_{Q}^{(h)}) (\boldsymbol{z}_j W_{K}^{(h)})^{\top}}{\sqrt{d_{\text{model}}/H}};\ \  a_{ij}^{(h)} = \frac{\text{exp}(e_{ij}^{(h)})}{\sum_{l=1}^{|Z|} \text{exp}(e_{il}^{(h)})}\\
 \boldsymbol{c}_i^{(h)} &= \sum_{j=1}^{|Z|}a_{ij}^{(h)} (\boldsymbol{z}_j W_{V}^{(h)});
  \boldsymbol{c}_i = \text{Concat}(\boldsymbol{c}_i^{(1)}, \cdots, \boldsymbol{c}_i^{(H)}) W_{O}
\end{align*}}where $1\leq h \leq H$, $W_O \in \mathbb{R}^{d_{\text{model}} \times d_{\text{model}}}$, and $W_{Q}^{(h)}, W_{K}^{(h)}, W_{V}^{(h)} \in \mathbb{R}^{d_{\text{model}} \times (d_{\text{model}}/H)}$. We can compute $\boldsymbol{c}_i$ for every $\boldsymbol{y}_i$ and get a transformed matrix $C \in \mathbb{R}^{|Y| \times d_{\text{model}}}$. The entire process is denoted as a mapping $\text{MultiHead}_\Theta$:
\begin{equation}
    C = \text{MultiHead}_\Theta(Y, Z) \label{eqn:mh}
\end{equation}


\noindent \textbf{Graph-based Multi-head Attention} To apply the multi-head attention on a graph, the graph adjacency matrix $\textbf{A}\in \mathbb{R}^{|Y| \times |Z|}$ is involved to mask nodes/tokens unrelated, where $\textbf{A}_{ij}\in\{0, 1\}$. Thus, $e_{ij}^{(h)}$ is changed as:
\begin{equation*}
e_{ij}^{(h)}=\left\{\begin{array}{ll}{\frac{(\boldsymbol{y}_i W_{Q}^{(h)}) (\boldsymbol{z}_j W_{K}^{(h)})^{\top}}{\sqrt{d_{\text{model}}/H}},} & {\text { if\ } \textbf{A}_{ij}=1} \\ {-\infty,} & {\text { otherwise }}\end{array}\right.
\end{equation*}
and Eqn. (\ref{eqn:mh}) is modified as:
\begin{equation}
    C = \text{GraphMultiHead}_\Theta(Y, Z, \textbf{A}) \label{eqn:gmh}
\end{equation}
Eqn. (\ref{eqn:mh}), can be treated as a special case of Eqn. (\ref{eqn:gmh}) that the graph is fully connected, i.e., $\textbf{A}=\boldsymbol{1}$.

\subsubsection{Context- and Schema-Aware Encoding}

Each layer of CSFN consists of internal and external attentions to incorporate different types of inputs. The hidden states of the schema graph $G$ at the $i$-the layer are updated as follows:
{
\begin{align*}
    \mathrm{I}_{\text{GG}} &= \text{\small GraphMultiHead}_{\Theta_{\text{GG}}}(\mathrm{H}_i^{G}, \mathrm{H}_i^{G}, \textbf{A}^G) \\
    \mathrm{E}_{\text{GX}} &= \text{\small MultiHead}_{\Theta_{\text{GX}}}(\mathrm{H}_i^{G}, \mathrm{H}_i^{X_t}) \\
    \mathrm{E}_{\text{GB}} &= \text{\small MultiHead}_{\Theta_{\text{GB}}}(\mathrm{H}_i^{G}, \mathrm{H}_i^{B_{t-1}}) \\
    \mathrm{C}_{{G}} &= \text{\small LayerNorm}(\mathrm{H}_i^{G} + \mathrm{I}_{\text{GG}} + \mathrm{E}_{\text{GX}} + \mathrm{E}_{\text{GB}}) \\
    \mathrm{H}_{i+1}^{G} &= \text{\small LayerNorm}(\mathrm{C}_{{G}} + \text{FFN}(\mathrm{C}_{{G}}))
\end{align*}}where $\textbf{A}^G$ is the adjacency matrix of the schema graph and $\text{LayerNorm}(.)$ is layer normalization function~\cite{ba2016layer}. $\text{FFN}(x)$ is a feed-forward network (FFN) function with two fully-connected layer and an ReLU activation in between, i.e., $\mathrm{FFN}(x)=\max \left(0, x W_{1}+b_{1}\right) W_{2}+b_{2}$.

Similarly, more details about updating $\mathrm{H}_{i}^{X_t}, \mathrm{H}_{i}^{B_{t-1}}$ are described in Appendix \ref{sec:appendix_fusion_network}.

The context and schema-aware encoding can also be simply implemented as the original transformer \cite{vaswani2017attention} with graph-based multi-head attentions.


\subsection{State Prediction}

The goal of state prediction is to produce the next dialogue state $B_t$, which is formulated as two stages: 1) We first apply a slot-gate classifier for each domain-slot node. The classifier makes a decision among \{\texttt{NONE}, \texttt{DONTCARE}, \texttt{PTR}\}, where \texttt{NONE} denotes that a domain-slot pair is not mentioned at this
turn, \texttt{DONTCARE} implies that the user can accept any values for this slot, and \texttt{PTR} represents that the slot should be processed with a value. 2) For domain-slot pairs tagged with \texttt{PTR}, we further introduced an RNN-based value decoder to generate token sequences of their values.

\subsubsection{Slot-gate Classification}

We utilize the final hidden vector of $j$-th domain-slot node in $G$ for the slot-gate classification, and the probability for the $j$-th domain-slot pair at the $t$-th turn is calculated as:
\begin{equation*}
    P_{tj}^{\text{gate}} = \text{softmax}(\text{FFN}(\mathrm{H}_{L,M+N+j}^{G}))
\end{equation*}
The loss for slot gate classification is 
\begin{equation*}
\mathcal{L}_{\text {gate}}=-\sum_{t=1}^{T} \sum_{j=1}^{J}\log (P_{t j}^\text{gate} \cdot (y_{tj}^\text{gate})^{\top})
\end{equation*}
where $y_{tj}^\text{gate}$ is the one-hot gate label for the $j$-th domain-slot pair at turn $t$.

\subsubsection{RNN-based Value Decoder}

After the slot-gate classification, there are $J'$ domain-slot pairs tagged with \texttt{PTR} class which indicates the domain-slot should take a real value. They are denoted as $\mathbb{C}_{t}=\{j|\text{argmax}(P_{tj}^{\text{gate}})=\texttt{PTR}\}$, and $J'=|\mathbb{C}_{t}|$. 

We use Gated Recurrent Unit (GRU) \cite{cho2014properties} decoder like \citet{wu2019transferable} and the soft copy mechanism \cite{see2017get} to get the final output distribution $P_{tj}^{\text{value}, k}$ over all candidate tokens at the $k$-th step. More details are illustrated in Appendix \ref{sec:appendix_rnn_value_decoder}. The loss function for value decoder is 
\begin{equation*}
    \mathcal{L}_{\text {value}}=-\sum_{t=1}^{T} \sum_{j\in \mathbb{C}_{t}}\sum_{k}\log (P_{tj}^{\text{value},k} \cdot(y_{tj}^{\text{value},k})^{\top})
\end{equation*}
where $y_{tj}^{\text{value},k}$ is the one-hot token label for the $j$-th domain-slot pair at $k$-th step.

During training process, the above modules can be jointly trained and optimized by the summations of different losses as:

\begin{equation*}
\mathcal{L}_{\text{total}}=\mathcal{L}_{\text{gate}}+\mathcal{L}_{\text{value}}
\end{equation*}

\section{Experiment}
\label{sec:exp}
\subsection{Datasets}
\label{subsec:exp_dataset}
We use MultiWOZ 2.0 \cite{budzianowski-etal-2018-multiwoz} and MultiWOZ 2.1 \cite{eric2019multiwoz} to evaluate our approach. MultiWOZ 2.0 is a task-oriented dataset of human-human written conversations spanning over seven domains, consists of 10348 multi-turn dialogues. MultiWOZ 2.1 is a revised version of MultiWOZ 2.0, which is re-annotated with a different set of inter-annotators and also canonicalized entity names. According to the work of \citet{eric2019multiwoz}, about $32\%$ of the state annotations is corrected so that the effect of noise is counteracted.

Note that \textit{hospital} and \textit{police} are excluded since they appear in training set with a very low frequency, and they do not even appear in the test set. To this end, five domains (\textit{restaurant}, \textit{train}, \textit{hotel}, \textit{taxi}, \textit{attraction}) are involved in the experiments with $17$ distinct slots and $30$ \textit{domain-slot} pairs. 

We follow similar data pre-processing procedures as~\citet{wu2019transferable} on both MultiWOZ 2.0 and 2.1.~\footnote{\url{https://github.com/budzianowski/multiwoz}} The resulting corpus includes 8,438 multi-turn dialogues in training set with an average of 13.5 turns per dialogue. Data statistics of MultiWOZ 2.1 is shown in Table \ref{tab:details}. The adjacency matrix $\textbf{A}^G$ of MultiWOZ 2.0 and 2.1 datasets is shown in Figure \ref{fig:adjacency_matrix} of Appendix, while domain-slot pairs are omitted due to space limitations.
\begin{table}[htp]
    \small
    \centering
    \begin{tabular}{|c|p{2.7cm}|p{0.5cm}p{0.5cm}p{0.5cm}|}
        \hline
        \textbf{Domain} & \textbf{Slots} & \textit{Train} & \textit{Valid} & \textit{Test}\\\hline
        \text{Restaurant}      & area, food, name, price range, book day, book people, book time &3813 & 438 & 437 \\\hline 
        \text{Hotel}           & area, internet, name, parking, price range, stars, type, book day, book people, book stay & 3381 & 416 & 394 \\\hline
        \text{Train}           & arrive by, day, departure, destination, leave at, book people & 3103 & 484 & 494 \\\hline
        \text{Taxi}            & arrive by, departure, destination, leave at & 1654 & 207 & 195 \\\hline
        \text{Attraction}      & area, name, type & 2717 & 401 & 395 \\\hline
        \multicolumn{2}{|c|}{\textit{Total}}        &    8438 & 1000  & 1000 \\\hline    
    \end{tabular}
    \caption{Data statistics of MultiWOZ2.1.}
    \label{tab:details}
\end{table}
 
 \begin{table*}[htbp]
    \small
    \centering
\begin{tabular}{c|l|c||cc}
\hline
\textbf{}  & \textbf{Models} & \textbf{BERT used} & \textbf{MultiWOZ 2.0} & \textbf{MultiWOZ 2.1} \\ \hline\hline
\multirow{9}{*}{\begin{tabular}[c]{@{}c@{}}predefined\\ ontology\end{tabular}} & HJST~\cite{eric2019multiwoz}* & \xmark & 38.40  & 35.55  \\ 
           & FJST~\cite{eric2019multiwoz}* & \xmark & 40.20  & 38.00  \\ 
           & SUMBT~\cite{lee2019sumbt}           & \cmark                 & 42.40  & -      \\ 
           & HyST~\cite{goel2019hyst}*            & \xmark & 42.33  & 38.10  \\ 
           & DS-DST~\cite{zhang2019find}          & \cmark                 & -      & 51.21  \\ 
           & DST-Picklist~\cite{zhang2019find}    & \cmark                 & -      & {53.30}          \\ 
           & DSTQA~\cite{zhou2019multi}           & \xmark & \textbf{51.44}          & 51.17  \\
           & SST~\cite{chen2020schema} & \xmark & 51.17          & \textbf{55.23}\\ \hline
\multirow{10}{*}{\begin{tabular}[c]{@{}c@{}}open-\\ vocabulary\end{tabular}}
          & DST-Span~\cite{zhang2019find}        & \cmark & -      & 40.39  \\ 
           & DST-Reader~\cite{gao-etal-2019-dialog}*      & \xmark & 39.41  & 36.40  \\ 
           & TRADE~\cite{wu2019transferable}*           & \xmark & 48.60  & 45.60  \\ 
           & COMER~\cite{ren-etal-2019-scalable}           & \cmark & 48.79  & -      \\ 
           & NADST~\cite{Le2020Non-Autoregressive}           & \xmark & 50.52  & 49.04      \\ 
           & SOM-DST~\cite{kim2019efficient}           & \cmark & 51.72  & {53.01}      \\ 
           \cline{2-5} 
           & CSFN-DST (ours) & \xmark &    49.59    & 50.81  \\ 
           & CSFN-DST + BERT (ours)  & \cmark                 & {51.57}          & {52.88}          \\
           & SOM-DST (our implementation)  & \cmark                 & 51.66          & 52.85          \\
           & SOM-DST + Schema Graph (ours)  & \cmark                 & \textbf{52.23}          & \textbf{53.19}          \\ \hline
\end{tabular}
    \caption{Joint goal accuracy (\%) on the test set of MultiWOZ 2.0 and 2.1. * indicates a result borrowed from \citet{eric2019multiwoz}. \cmark means that a BERT model \cite{devlin2018bert} with contextualized word embeddings is utilized.} 
    \label{tab:main_res}
\end{table*}

\subsection{Experiment Settings}
\label{subsec:exp_set}
We set the hidden size of CSFN, $d_{\text{model}}$, as 400 with 4 heads.
Following \citet{wu2019transferable}, the token embeddings with 400 dimensions are initialized by concatenating Glove embeddings \cite{pennington-etal-2014-glove} and character embeddings \cite{hashimoto-etal-2017-joint}. We do a grid search over $\{4,5,6,7,8\}$ for the layer number of CSFN on the validation set. We use a batch size of 32. The DST model is trained using ADAM \cite{kingma2014adam} with the learning rate of 1e-4. During training, we use the ground truth of the previous dialogue state and the ground truth value tokens. In the inference, the predicted dialogue state of the last turn is applied, and we use a greedy search strategy in the decoding process of the value decoder. 


\subsection{Baseline Models}
\label{subsec:exp_baseline}

We make a comparison with the following existing models, which are either predefined ontology-based DSTs or open-vocabulary based DSTs. Predefined ontology-based DSTs have the advantage of access to the known candidate set of each slot, while these approaches may not be applicable in the real scenario.


\noindent\textbf{FJST}~\cite{eric2019multiwoz}: It exploits a bidirectional LSTM network to encode the dialog history and a separate FFN to predict the value for each slot.

\noindent\textbf{HJST}~\cite{eric2019multiwoz}: It encodes the dialogue history using an LSTM like FJST, but utilizes a hierarchical network.

\noindent\textbf{SUMBT}~\cite{lee2019sumbt}: It exploits BERT~\cite{devlin2018bert} as the encoder for the dialogue context and slot-value pairs. After that, it scores every candidate slot-value pair with the dialogue context by using a distance measure.


\noindent\textbf{HyST}~\cite{goel2019hyst}: It is a hybrid approach based on hierarchical RNNs, which incorporates both a predefined ontology-based setting and an open-vocabulary setting.

\noindent\textbf{DST-Reader}~\cite{gao-etal-2019-dialog}: It models the DST from the perspective of text reading comprehensions, and get start and end positions of the corresponding text span in the dialogue context.

\noindent\textbf{DST-Span}~\cite{zhang2019find}: It treats all domain-slot pairs as span-based slots like DST-Reader, and applies a BERT as the encoder.


\noindent\textbf{DST-Picklist}~\cite{zhang2019find}: It defines picklist-based slots for classification similarly to SUMBT and applies a pre-trained BERT for the encoder. It relies on a predefined ontology.

\noindent\textbf{DS-DST}~\cite{zhang2019find}: Similar to HyST, it is a hybrid system of DS-Span and DS-Picklist.  



\noindent\textbf{DSTQA}~\cite{zhou2019multi}: It models multi-domain DST as a question answering problem, and generates a question asking for the value of each domain-slot pair. It heavily relies on a predefined ontology, i.e., the candidate set for each slot is known, except for five time-related slots.

\noindent\textbf{TRADE}~\cite{wu2019transferable}: It contains a slot gate module for slots classification and a pointer generator for dialogue state generation. 

\noindent\textbf{COMER}~\cite{ren-etal-2019-scalable}: It uses a hierarchical decoder to generate the current dialogue state itself as the target sequence.  

\noindent\textbf{NADST}~\cite{Le2020Non-Autoregressive}: It uses a non-autoregressive decoding scheme to generate the current dialogue state.

\noindent\textbf{SST}~\cite{chen2020schema}: It utilizes a graph attention matching network to fuse information from utterances and schema graphs, and a recurrent graph attention network to control state updating. However, it heavily relies on a predefined ontology.

\noindent\textbf{SOM-DST}~\cite{kim2019efficient}: It uses a BERT to jointly encode the previous state, the previous and current dialogue utterances. An RNN-decoder is also applied to generate values for slots that need to be updated in the open-vocabulary setting.

\subsection{Main Results}

Joint goal accuracy is the evaluation metric in our experiments, which is represented as the ratio of turns whose predicted dialogue states are entirely consistent with the ground truth in the test set. Table \ref{tab:main_res} illustrates that the joint goal accuracy of CSFN-DST and other baselines on the test set of MultiWOZ 2.0 and MultiWOZ 2.1 datasets. 

As shown in the table, our proposed CSFN-DST can outperform other models except for SOM-DST. By combining our schema graphs with SOM-DST, we can achieve state-of-the-art performances on both MultiWOZ 2.0 and 2.1 in the open-vocabulary setting. Additionally, our method using BERT (\texttt{Bert-base-uncased}) can obtain very competitive performance with the best systems in the predefined ontology-based setting. When a BERT is exploited, we initialize all parameters of CSFN with the BERT encoder's and initialize the token/position embeddings with the BERT's.


\subsection{Analysis}

In this subsection, we will conduct some ablation studies to figure out the potential factors for the improvement of our method. (Additional experiments and results are reported in Appendix \ref{sec:appendix_addtional_results}, case study is shown in Appendix \ref{sec:appendix_case_study}.)

\subsubsection{Effect of context information}

Context information consists of the previous dialogue state or the current dialogue utterance, which are definitely key for the encoder. It would be interesting to know whether the two kinds of context information are also essential for the RNN-based value decoder. As shown in Table \ref{tab:context_information}, we choose to omit the top hidden states of the previous dialogue state ($\mathrm{H}_{L}^{B_{t-1}}$) or the current utterance ($\mathrm{H}_{L}^{X_t}$) in the RNN-based value decoder. The results show both of them are crucial for generating real values.

\textbf{Do we need more context?} Only the current dialogue utterance is utilized in our model, which would be more efficient than the previous methods involving all the preceding dialogue utterance. However, we want to ask whether the performance will be improved when more context is used. In Table \ref{tab:context_information}, it shows that incorporating the previous dialogue utterance $X_{t-1}$ gives no improvement, which implies that jointly encoding the current utterance and the previous dialogue state is effective as well as efficient.

\begin{table}[t]
    \small
    \centering
    \begin{tabular}{|l|c|}\hline
        Models & Joint Acc. (\%) \\\hline
         CSFN-DST  & 50.81   \\\hline 
         \quad (-) Omit $\mathrm{H}_{L}^{B_{t-1}}$ in the decoder & 48.66  \\
         \quad (-) Omit $\mathrm{H}_{L}^{X_t}$ in the decoder     & 48.45  \\
         \quad (+) The previous utterance $X_{t-1}$          & 50.75      \\\hline
    \end{tabular}
    \caption{Ablation studies for context information on MultiWOZ 2.1.}
    \label{tab:context_information}
\end{table}

\subsubsection{Effect of the schema graph}


\begin{table}[]
    \small
    \centering
    \begin{tabular}{|l|c|c|}\hline
    \multirow{2}{*}{\begin{tabular}[c]{@{}l@{}}Models\end{tabular}} & \multicolumn{2}{c|}{BERT used} \\ \cline{2-3}
         & \xmark & \cmark \\\hline
         CSFN-DST                           & 50.81   & 52.88            \\\hline 
         \ \ (-) No schema graph, $\textbf{A}^G=\boldsymbol{1}$               & 49.93   &    52.50        \\
         \ \ (-) No schema graph, $\textbf{A}^G=\boldsymbol{I}$              & 49.52    &    52.46      \\\hline
         \ \ (+) Ground truth of the previous state          &  78.73 & 80.35          \\\hline
         \ \ (+) Ground truth slot-gate classifi.          & 77.31  &  80.66         \\\hline
         \ \ (+) Ground truth value generation          & 56.50  & 59.12             \\\hline
    \end{tabular}
    \caption{Joint goal accuracy(\%) of ablation studies on MultiWOZ 2.1.}
    \label{tab:ablation}
\end{table}



In CSFN-DST, the schema graph with domain-slot relations is exploited. To check the effectiveness of the schema graph used, we remove knowledge-aware domain-slot relations by replacing the adjacency matrix $\textbf{A}^G$ as a fully connected one $\boldsymbol{1}$ or node-independent one $\boldsymbol{I}$. Results in Table \ref{tab:ablation} show that joint goal accuracies of models without the schema graph are decreased similarly when BERT is either used or not. 

To reveal why the schema graph with domain-slot relations is essential for joint accuracy, we further make analysis on domain-specific and turn-specific results. As shown in Table \ref{tab:domain_joint}, the schema graph can benefit almost all domains except for \emph{Attaction (Attr.)}. As illustrated in Table \ref{tab:details}, the \emph{Attaction} domain contains only three slots, which should be much simpler than the other domains. Therefore, we may say that the schema graph can help complicated domains. 

The turn-specific results are shown in Table \ref{tab:different_turns}, where joint goal accuracies over different dialogue turns are calculated. From the table, we can see that data proportion of larger turn number becomes smaller while the larger turn number refers to more challenging conversation. From the results of the table, we can find the schema graph can make improvements over most dialogue turns.

\begin{table}[]
    \small
    \centering
    \begin{tabular}{|c||c|c|c|c|c|}\hline
         Models & Attr. & Hotel & Rest. & Taxi & Train \\\hline\hline 
         CSFN-DST & 64.78 & \textbf{46.29} & \textbf{64.64} & \textbf{47.35} &  \textbf{69.79} \\
         \ \ (-) No SG & \textbf{65.97}  &  45.48 & 62.94 & 46.42 &  67.58 \\\hline
    \end{tabular}
    \caption{Domain-specific joint accuracy on MultiWOZ 2.1. SG means Schema Graph.}
    \label{tab:domain_joint}
\end{table}



\begin{table}[]
    \small
    \centering
    \begin{tabular}{|c|c||l|l|}
    \hline
         \text{Turn} & Proportion (\%) & w/ SG & w/o SG \\\hline\hline
         \text{1} & 13.6 & 89.39  &  88.19 ($-1.20$) \\
         \text{2} & 13.6 & 73.87  &  72.87 ($-1.00$) \\
         \text{3} & 13.4 & 58.69  &  57.78 ($-0.91$) \\
         \text{4} & 12.8 & 51.96  &  50.80 ($-1.16$) \\
         \text{5} & 11.9 & 41.01  &  39.63 ($-1.38$) \\
         \text{6} & 10.7 & 34.51  &  35.15 ($+0.64$) \\
         \text{7} & 9.1 & 27.91  &  29.55 ($+1.64$) \\
         \text{8} & 6.3 & 24.73  &  23.23 ($-1.50$) \\
         \text{9} & 4.0 & 20.55  &  19.18 ($-1.37$) \\
         \text{10} & 2.3 & 16.37  &  12.28 ($-4.09$) \\
         \text{11} & 1.3 & 12.63  &  8.42 ($-4.21$) \\
         \text{12} & 0.6 & 12.77  &  8.51 ($-4.26$) \\
         \text{$>12$}& 0.4 & 9.09 & 0.00 ($-9.09$) \\\hline
         \text{all}& 100 & 50.81  &  49.93      \\\hline
    \end{tabular}
    \caption{Joint accuracies over different dialogue turns on MultiWOZ 2.1. It shows the impact of using schema graph on our proposed CSFN-DST.}
    \label{tab:different_turns}
\end{table}


\subsubsection{Oracle experiments}

The predicted dialogue state at the last turn is utilized in the inference stage, which is mismatched with the training stage. An oracle experiment is conducted to show the impact of training-inference mismatching, where ground truth of the previous dialogue state is fed into CSFN-DST. The results in Table \ref{tab:ablation} show that joint accuracy can be nearly $80\%$ with ground truth of the previous dialogue state. Other oracle experiments with ground truth slot-gate classification and ground truth value generation are also conducted, as shown in Table \ref{tab:ablation}.


\subsubsection{Slot-gate classification}

We conduct experiments to evaluate our model performance on the slot-gate classification task. Table \ref{tab:gate} shows F1 scores of the three slot gates, i.e., \{\texttt{NONE}, \texttt{DONTCARE}, \texttt{PTR}\}. It seems that the pre-trained BERT model helps a lot in detecting slots of which the user doesn't care about values. The F1 score of \texttt{DONTCARE} is much lower than the others', which implies that detecting \texttt{DONTCARE} is a much challenging sub-task.

\begin{table}[h]
    \small
    \centering
    \begin{tabular}{c|c|c}\hline
        \textbf{Gate} & \text{CSFN-DST}   & \text{CSFN-DST} + BERT \\\hline\hline
        \texttt{NONE} & 99.18 & 99.19          \\\hline
    \texttt{DONTCARE} & 72.50 & 75.96      \\\hline
         \texttt{PTR} & 97.66 & 98.05       \\\hline
    \end{tabular}
    \caption{Slot-gate F1 scores on MultiWOZ 2.1.}
    \label{tab:gate}
\end{table}

\subsection{Reproducibility}

We run our models on GeForce GTX 2080 Ti Graphics Cards, and the average training time for each epoch and number of parameters in each model are provided in Table \ref{tab:reproducibility}. If BERT is exploited, we accumulate the gradients with 4 steps for a minibatch of data samples (i.e., $32/4=8$ samples for each step), due to the limitation of GPU memory. As mentioned in Section 5.4, joint goal accuracy is the evaluation metric used in our experiments, and we follow the computing script provided in TRADE-DST~\footnote{\url{https://github.com/jasonwu0731/trade-dst}}.

\begin{table}[htp]
    \small
    \centering
    \begin{tabular}{|c|c|c|}
        \hline
        \textbf{Method} & \textbf{Time per Batch} & \textbf{\# Parameters} \\
        \hline
        CSFN-DST & 350ms & 63M \\
        CSFN-DST + BERT & 840ms & 115M \\
        SOM-DST + SG & 1160ms & 115M \\
        \hline    
    \end{tabular}
    \caption{Runtime and mode size of our methods.}
    \label{tab:reproducibility}
\end{table}

\subsection{Discussion}

The main contributions of this work may focus on exploiting the schema graph with graph-based attention networks. Slot-relations are also utilized in DSTQA~\cite{zhou2019multi}. However, DSTQA uses a dynamically-evolving knowledge graph for the dialogue context, and we use a static schema graph. We absorb the dialogue context by using the previous (predicted) dialogue state as another input. We believe that the two different usages of the slot relation graph can be complementary. Moreover, these two methods are different in value prediction that DSTQA exploits a hybrid of value classifier and span prediction layer, which relies on a predefined ontology. 

SOM-DST~\cite{kim2019efficient} is very similar to our proposed CSFN-DST with BERT. The main difference between SOM-DST and CSFN-DST is how to exploit the previous dialogue state. For the previous dialogue state, SOM-DST considers all domain-slot pairs and their values (if a domain-slot pair contains an empty value, a special token NONE is used), while CSFN-DST only considers the domain-slot pairs with a non-empty value. Thus, SOM-DST knows which domain-slot pairs are empty and would like to be filled with a value. We think that it is the strength of SOM-DST. However, we choose to omit the domain-slot pairs with an empty value for a lower computation burden, which is proved in Table \ref{tab:reproducibility}. As shown in the last two rows of Table \ref{tab:main_res}, the schema graph can also improve SOM-DST, which achieves $52.23\%$ and $53.19\%$ joint accuracies on MultiWOZ 2.0 and 2.1, respectively. Appendix \ref{sec:som_dst_with_sg} shows how to exploit schema graph in SOM-DST.


\section{Conclusion and Future Work}

We introduce a multi-domain dialogue state tracker with context and schema fusion networks, which involves slot relations and learns deep representations for each domain-slot pair dependently. Slots from different domains and their relations are organized as a schema graph. Our approach outperforms strong baselines on both MultiWOZ 2.0 and 2.1 benchmarks. Ablation studies also show that the effectiveness of the schema graph.  

It will be a future work to incorporate relations among dialogue states, utterances and domain schemata. To further mitigate the data sparsity problem of multi-domain DST, it would be also interesting to incorporate data augmentations~\cite{zhao2019data} and semi-supervised learnings~\cite{lan2018semi,cao2019semantic}. 


\section*{Acknowledgments}

We thank the anonymous reviewers for their thoughtful comments. This work has been supported by Shanghai Jiao Tong University Scientific and Technological Innovation Funds (YG2020YQ01) and No. SKLMCPTS2020003 Project.


\bibliography{emnlp2020}
\bibliographystyle{acl_natbib}

\appendix

\section{Context- and Schema-Aware Encoding}
\label{sec:appendix_fusion_network}
Besides the hidden states $\mathrm{H}_{i}^G$ of the schema graph $G$, we show the details of updating $\mathrm{H}_{i}^{X_t}, \mathrm{H}_{i}^{B_{t-1}}$ in the $i$-th layer of CSFN:
{\small
\begin{equation*}
    \mathrm{H}_{i+1}^{G}, \mathrm{H}_{i+1}^{X_t}, \mathrm{H}_{i+1}^{B_{t-1}} = \text{\small CSFNLayer}_i(\mathrm{H}_{i}^{G}, \mathrm{H}_{i}^{X_t}, \mathrm{H}_{i}^{B_{t-1}})
\end{equation*}}

The hidden states of the dialogue utterance $X_t$ at the $i$-the layer are updated as follows:
\begin{align*}
    \mathrm{I}_{\text{XX}} &= \text{\small MultiHead}_{\Theta_{\text{XX}}}(\mathrm{H}_i^{X_t}, \mathrm{H}_i^{X_t}) \\
    \mathrm{E}_{\text{XB}} &= \text{\small MultiHead}_{\Theta_{\text{XB}}}(\mathrm{H}_i^{X_t}, \mathrm{H}_i^{B_{t-1}}) \\
    \mathrm{E}_{\text{XG}} &= \text{\small MultiHead}_{\Theta_{\text{XG}}}(\mathrm{H}_i^{X_t}, \mathrm{H}_i^{G}) \\
    \mathrm{C}_X &= \text{\small LayerNorm}(\mathrm{H}_i^{X_t} + \mathrm{I}_{\text{XX}} + \mathrm{E}_{\text{XB}} + \mathrm{E}_{\text{XG}}) \\
    \mathrm{H}_{i+1}^{X_t} &= \text{\small LayerNorm}(\mathrm{C}_X + \text{FFN}(\mathrm{C}_X))
\end{align*}
where $\mathrm{I}_{\text{XX}}$ contains internal attention vectors, $\mathrm{E}_{\text{XB}}$ and $\mathrm{E}_{\text{XG}}$ are external attention vectors.

The hidden states of the previous dialogue state $B_{t-1}$ at the $i$-the layer are updated as follows:
\begin{align*}
    \mathrm{I}_{\text{BB}} &= \text{\small GraphMultiHead}_{\Theta_{\text{BB}}}(\mathrm{H}_i^{B_{t-1}}, \mathrm{H}_i^{B_{t-1}}, \textbf{A}^{B_{t-1}}) \\
    \mathrm{E}_{\text{BX}} &= \text{\small MultiHead}_{\Theta_{\text{BX}}}(\mathrm{H}_i^{B_{t-1}}, \mathrm{H}_i^{X_t}) \\
    \mathrm{E}_{\text{BG}} &= \text{\small MultiHead}_{\Theta_{\text{BG}}}(\mathrm{H}_i^{B_{t-1}}, \mathrm{H}_i^{G}) \\
    \mathrm{C}_B &= \text{\small LayerNorm}(\mathrm{H}_i^{B_{t-1}} + \mathrm{I}_{\text{BB}} + \mathrm{E}_{\text{BX}} + \mathrm{E}_{\text{BG}}) \\
    \mathrm{H}_{i+1}^{B_{t-1}} &= \text{\small LayerNorm}(\mathrm{C}_B + \text{FFN}(\mathrm{C}_B))
\end{align*}
where $\textbf{A}^{B_{t-1}}$ is the adjacency matrix of the previous dialogue state. The adjacency matrix indicates that each triplets in ${B_{t-1}}$ is separated, while tokens in a same triplet are connected with each other. The \texttt{[CLS]} token is connected with all triplets, serving as a transit node.

\section{RNN-based Value Decoder}
\label{sec:appendix_rnn_value_decoder}

After the slot-gate classification, there are $J'$ domain-slot pairs tagged with \texttt{PTR} class which indicates the domain-slot should take a real value. They are denoted as $\mathbb{C}_{t}=\{j|\text{argmax}(P_{tj}^{\text{gate}})=\texttt{PTR}\}$, and $J'=|\mathbb{C}_{t}|$. 

We use Gated Recurrent Unit (GRU) \cite{cho2014properties} decoder like \citet{wu2019transferable} and \citet{see2017get}. The hidden state $g_{tj}^k \in \mathbb{R}^{1\times d_{\text{model}}}$ is recursively updated by taking a word embedding $e_{tj}^k$ as the input until
\texttt{[EOS]} token is generated:
\begin{equation*}
    g_{tj}^{k}=\operatorname{GRU}(g_{tj}^{k-1}, e_{tj}^{k})
\end{equation*}
GRU is initialized with 
\begin{equation*}
g_{tj}^0=\mathrm{H}_{L,0}^{X_t} + \mathrm{H}_{L,0}^{B_{t-1}}
\end{equation*}
and $e_{tj}^0=\mathrm{H}_{L,M+N+j}^{G}$.

The value generator transforms the hidden state to the probability distribution over the token vocabulary at the $k$-th step, which consists of two parts: 1) distribution over all input tokens, 2) distribution over the input vocabulary. The first part is computed as
\begin{equation*}
\begin{aligned}
P_{tj}^{\text{ctx}, k} &=\text{softmax}(\textsc{Att}(g_{tj}^{k}, [\mathrm{H}_{L}^{X_t};\mathrm{H}_{L}^{B_{t-1}}])) 
\end{aligned}
\end{equation*}
where $P_{tj}^{\text{ctx}, k} \in \mathbb{R}^{1\times (|X_t|+|B_{t-1}|)}$, and  $\textsc{Att}(.,.)$ is a function to get attention weights~\cite{bahdanau2014neural} with more details shown in Appendix \ref{sec:appendix_attention_weight}. The second part is calculated as
\begin{equation*}
\begin{aligned}
c_{tj}^{k}&=P_{tj}^{\text{ctx}, k} [\mathrm{H}_{L}^{X_t};\mathrm{H}_{L}^{B_{t-1}}]\\
P_{tj}^{\text{vocab}, k} &=\text{softmax}([g_{tj}^{k};c_{tj}^{k}]W_{\text{proj}} E^{\top})
\end{aligned}
\end{equation*}
where $P_{tj}^{\text{vocab}, k} \in \mathbb{R}^{1\times d_{\text{vocab}}}$, $c_{tj}^{k} \in \mathbb{R}^{1\times d_{\text{model}}}$ is a context vector, $W_\text{proj} \in \mathbb{R}^{2d_{\text{model}}\times d_{\text{model}}}$ is a trainable parameter, and $E \in \mathbb{R}^{d_{\text{vocab}}\times d_{\text{model}}}$ is the token embedding matrix shared across the encoder and the decoder.

We use the soft copy mechanism \cite{see2017get} to get the final output distribution over all candidate tokens:
\begin{align*}
P_{tj}^{\text{value}, k}&=p_\text{gen} P_{tj}^{\text{vocab}, k}+(1-p_\text{gen}) P_{tj}^{\text{ctx}, k}\\
p_\text{gen}&=\text{sigmoid}([g_{tj}^{k} ; e_{tj}^{k} ; c_{tj}^{k}]W_{\text{gen}})
\end{align*}
where $W_\text{gen} \in \mathbb{R}^{3d_{\text{model}}\times 1}$ is a trainable parameter.

The loss function for value decoder is 
\begin{equation*}
    \mathcal{L}_{\text {value}}=-\sum_{t=1}^{T} \sum_{j\in \mathbb{C}_{t}}\sum_{k}\log (P_{tj}^{\text{value},k} \cdot(y_{tj}^{\text{value},k})^{\top})
\end{equation*}
where $y_{tj}^{\text{value},k}$ is the one-hot token label for the $j$-th domain-slot pair at $k$-th step.

\subsection{Attention Weights}
\label{sec:appendix_attention_weight}

For attention mechanism for computing $P_{tj}^{\text{ctx}, k}$ in the RNN-based value decoder, we follow~\citet{bahdanau2014neural} and define the $\textsc{Att}(.,.)$ function as

\begin{align*}
    u_i=&\text{tanh}(\textbf{x}\textbf{W}_1^{\text{att}}+\textbf{h}_i\textbf{W}_2^{\text{att}}+\textbf{b}^{\text{att}})\textbf{v}^{\top} \\
    a_i=&\frac{\text{exp}(u_i)}{\sum_{j=1}^S\text{exp}(u_j)}\\
    \textbf{a}=&\{a_1,\cdots,a_S\}=\textsc{Att}(\textbf{x}, \text{H})
\end{align*}
where $\textbf{x} \in \mathbb{R}^{1\times d}$, $\text{H} \in \mathbb{R}^{S\times d}$, $\textbf{W}_1^{\text{att}} \in \mathbb{R}^{d \times d}$, $\textbf{W}_2^{\text{att}} \in \mathbb{R}^{d \times d}$, $\textbf{b}^{\text{att}} \in \mathbb{R}^{1\times d}$, $\textbf{v} \in \mathbb{R}^{1\times d}$, and $\textbf{h}_i$ is the $i$-th row vector of $\text{H}$. Therefore, $\textsc{Att}(\textbf{x}, \text{H})$ returns an attention distribution of $\textbf{x}$ over $\text{H}$.


\begin{figure*}[t]
    \centering
    \includegraphics[width=0.9\linewidth]{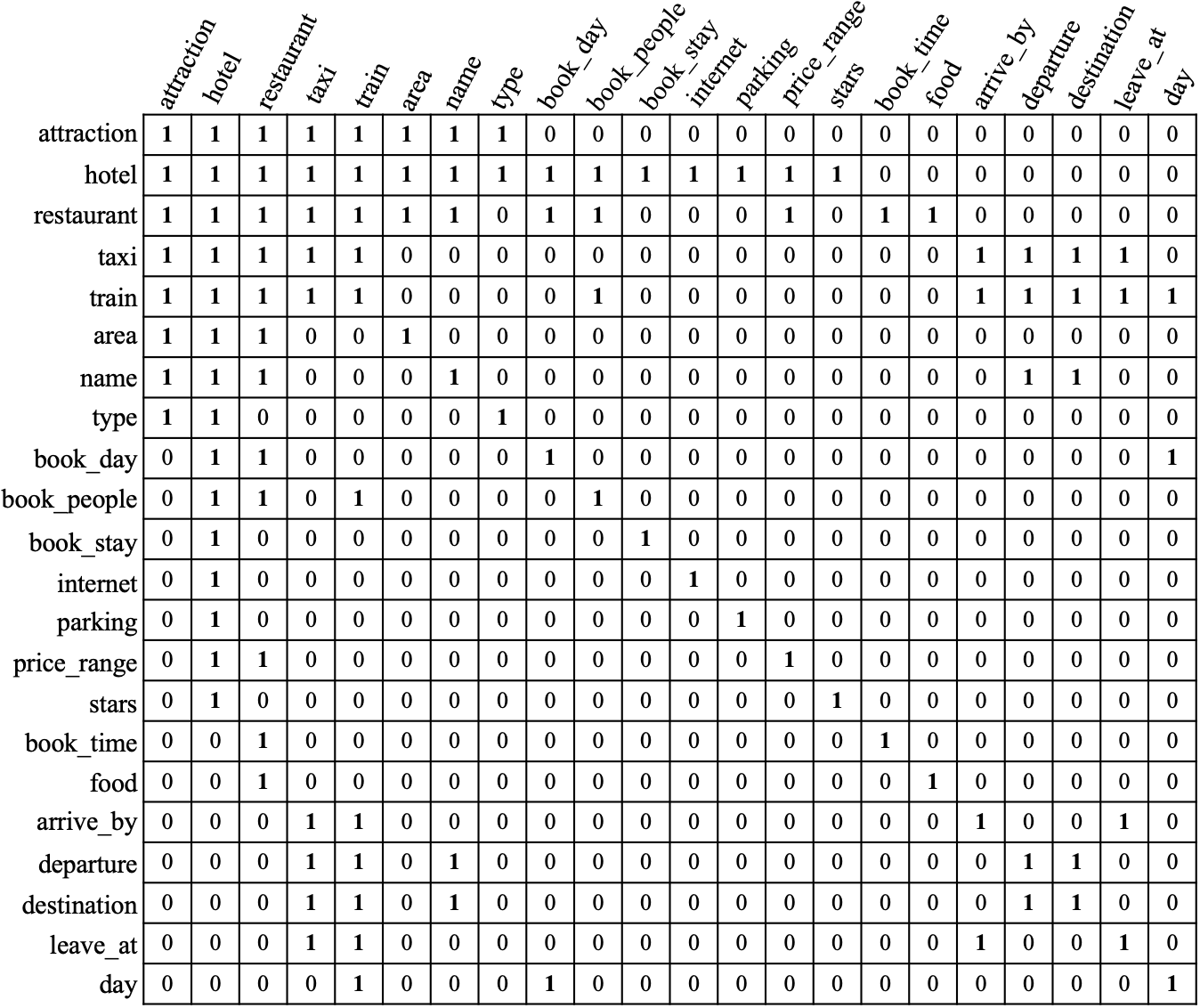}
    \caption{Adjacency matrix $\textbf{A}^G$ of MultiWOZ 2.0 and 2.1 datasets. It contains only domain and slot nodes, while domain-slot paris are omitted due to space limitations. The first five items are domains (``attraction, hotel, restaurant, taxi, train''), and the rest are slots.}
    \label{fig:adjacency_matrix}
\end{figure*}


\section{Additional Results}
\label{sec:appendix_addtional_results}

\noindent \textbf{Domain-specific Results} Domain-specific accuracy is the joint goal accuracy measured on a subset of the predicted dialogue state, which only contains the slots belong to a domain. From the results of Table \ref{tab:domain_specific}, we can find BERT can make improvements on all domains, and especially the improvement on \emph{Taxi} domain is the largest.

\begin{table}[h]
    \small
    \centering
    \begin{tabular}{|c||c|c|}\hline
         Domain & CSFN-DST & CSFN-DST + BERT \\\hline\hline 
         \text{Attraction}  & 64.78  & 67.82      \\\hline
         \text{Hotel}       & \text{46.29} &  48.80      \\\hline
         \text{Restaurant}  & \text{64.64} & 65.23        \\\hline
         \text{Taxi}        & \text{47.35} & 53.58       \\\hline
         \text{Train}       & \text{69.79} &  70.91       \\\hline
    \end{tabular}
    \caption{Domain-specific joint accuracy on MultiWOZ 2.1.}
    \label{tab:domain_specific}
\end{table}

\noindent \textbf{Slot-specific Results} Slot-specific F1 score is measured for predicting slot-value pairs of the corresponding slot. Table \ref{tab:slot_specific} shows slot-specific F1 scores of CSFN-DST without the schema graph, CSFN-DST and CSFN-DST with BERT on the test set of MultiWOZ 2.1.

\begin{table*}[h]
    \centering
    \begin{tabular}{|c||c|c|c|}\hline
         \textbf{Domain-slot} & CSFN-DST (no SG) & CSFN-DST & CSFN-DST + BERT \\\hline\hline
        attraction-area & 91.67 & 91.92 & \textbf{92.81} \\
        attraction-name & 78.28 & 77.77 & \textbf{79.55} \\
        attraction-type & 90.95 & 90.89 & \textbf{91.97} \\
        hotel-area & 84.59 & 84.21 & \textbf{84.86} \\
        hotel-book\_day & 97.16 & \textbf{97.79} & 97.03 \\
        hotel-book\_people & 95.43 & 96.14 & \textbf{97.35} \\
        hotel-book\_stay & 96.04 & \textbf{97.00} & 96.98 \\
        hotel-internet & 86.79 & \textbf{89.98} & 86.58 \\
        hotel-name & 82.97 & 83.11 & \textbf{84.61} \\
        hotel-parking & 86.68 & \textbf{87.66} & 86.07 \\
        hotel-price\_range & 89.09 & 90.10 & \textbf{92.56} \\
        hotel-stars & 91.51 & \textbf{93.49} & 93.34 \\
        hotel-type & 77.58 & 77.87 & \textbf{82.12} \\
        restaurant-area & 93.73 & \textbf{94.27} & 94.09 \\
        restaurant-book\_day & \textbf{97.75} & 97.66 & 97.42 \\
        restaurant-book\_people & 96.79 & 96.67 & \textbf{97.84} \\
        restaurant-book\_time & 92.43 & 91.60 & \textbf{94.29} \\
        restaurant-food & \textbf{94.90} & 94.18 & 94.48 \\
        restaurant-name & 80.72 & \textbf{81.39} & 80.59 \\
        restaurant-price\_range & 93.47 & 94.28 & \textbf{94.49} \\
        taxi-arrive\_by & 78.81 & 81.09 & \textbf{86.08} \\
        taxi-departure & 73.15 & 71.39 & \textbf{75.15} \\
        taxi-destination & 73.79 & 78.06 & \textbf{79.83} \\
        taxi-leave\_at & 77.29 & 80.13 & \textbf{88.06} \\
        train-arrive\_by & 86.43 & 87.56 & \textbf{88.77} \\
        train-book\_people & 89.59 & 91.41 & \textbf{92.33} \\
        train-day & 98.44 & 98.41 & \textbf{98.52} \\
        train-departure & 95.91 & \textbf{96.52} & 96.22 \\
        train-destination & \textbf{97.08} & 97.06 & 96.13 \\
        train-leave\_at & 69.97 & 70.97 & \textbf{74.50} \\
        \hline
        Joint Acc. overall & 49.93 & 50.81 & 52.88 \\ \hline
    \end{tabular}
    \caption{Slot-specific F1 scores on MultiWOZ 2.1. SG means Schema Graph. The results in bold black are the best slot F1 scores.}
    \label{tab:slot_specific}
\end{table*}

\section{Case Study}
\label{sec:appendix_case_study}
We also conduct case study on the test set of MultiWOZ 2.1, and four cases are shown in Table~\ref{tab:case_study}. From the first three cases, we can see the schema graph can copy values from related slots in the memory (i.e., the previous dialogue state). In the case C1, the model makes the accurate reference of the phrase ``\emph{whole group}'' through the
context, and the value of \emph{restaurant-book\_people} is copied as the value of \emph{train-book\_people}. We can also see a failed case (C4). It is too complicated to inference the \emph{departure} and \emph{destination} by a word ``\emph{commute}''.

\section{SOM-DST with Schema Graph}
\label{sec:som_dst_with_sg}

For SOM-DST~\cite{kim2019efficient}, the input tokens to the state operation predictor are the concatenation of the previous turn dialog utterances, the current turn dialog utterances, and the previous turn dialog state:
$$X_{t}=[\texttt{CLS}] \oplus D_{t-1} \oplus D_{t} \oplus B_{t-1},$$
where $ D_{t-1}$ and $D_{t}$ are the last and current utterances, respectively. The dialogue state $B_{t}$ is denoted as $B_{t}=B_{t}^{1} \oplus \ldots \oplus B_{t}^{J}$, where $B_{t}^{j}=[\texttt{SLOT}]^{j} \oplus S^{j} \oplus-\oplus V_{t}^{j}$ is the representation of the $j$-th slot-value pair. To incorporate the schema graph, we exploit the special token $[\texttt{SLOT}]^{j}$ to replace the domain-slot node $o_j$ in the schema graph ($j=1,\cdots,J$). Then, domain and slot nodes $G'=d_1 \oplus \cdots \oplus d_M \oplus s_1 \oplus \cdots \oplus s_N$ are concatenated into $X_t$, i.e.,
$$X_{t}=[\texttt{CLS}] \oplus D_{t-1} \oplus D_{t} \oplus B_{t-1} \oplus G',$$
where the relations among domain, slot and domain-slot nodes are also considered in attention masks of BERT.

\begin{table*}[h]
    \small
    \centering
    \begin{tabular}{l|ll}
        \hline
       \multirow{17}{*}{\textbf{C1}} & \textbf{Previous DS:} & \begin{tabular}[c]{@{}l@{}}(restaurant-book\_day, friday), (restaurant-book\_people, \textbf{8}), (restaurant-book\_time, 10:15),\\ (restaurant-name, restaurant 2 two), (train-leave\_at, 12:15),\\ (train-destination, peterborough), (train-day, saturday), (train-departure, cambridge)\end{tabular}\\
       & \textbf{System:} & How about train tr3934? It leaves at 12:34 \& arrives at 13:24. Travel time is 50 minutes. \\
        & \textbf{Human:} & That sounds fine. Can I get tickets for my \textbf{whole group} please?\\
        \cline{2-3}
        & \textbf{Gold DS:} & \begin{tabular}[c]{@{}l@{}}(restaurant-name, restaurant 2 two), (restaurant-book\_day, friday),\\ (restaurant-book\_people, \text{8}), (restaurant-book\_time, 10:15), (train-departure, cambridge),\\ (train-leave\_at, 12:15), (train-day, saturday), (train-destination, peterborough),\\ (train-book\_people, \textbf{8})\end{tabular}  \\
        & \textbf{CSFN-DST (no SG):} & \begin{tabular}[c]{@{}l@{}}(restaurant-name, restaurant 2 two), (restaurant-book\_day, friday),\\ (restaurant-book\_people, \text{8}), (restaurant-book\_time, 10:15), (train-departure, cambridge),\\ (train-leave\_at, 12:15), (train-day, saturday), (train-destination, peterborough),\\ (train-book\_people, \textbf{1})\end{tabular} \\
        & \textbf{CSFN-DST:} & \begin{tabular}[c]{@{}l@{}}(restaurant-name, restaurant 2 two), (restaurant-book\_day, friday),\\ (restaurant-book\_people, \text{8}), (restaurant-book\_time, 10:15), (train-departure, cambridge),\\ (train-leave\_at, 12:15), (train-day, saturday), (train-destination, peterborough),\\ (train-book\_people, \textbf{8})\end{tabular} \\
        \hline
        
       \multirow{17}{*}{\textbf{C2}} & \textbf{Previous DS:} & \begin{tabular}[c]{@{}l@{}}(hotel-area, west), (hotel-price\_range, cheap), (hotel-type, guest house),\\ (hotel-internet, yes), (hotel-name, warkworth house), (restaurant-area, centre),\\ (restaurant-food, italian), (restaurant-price\_range, cheap), (restaurant-name, \textbf{ask})\end{tabular}\\
       & \textbf{System:} & 01223364917 is the phone number. 12 bridge street city centre, cb21uf is the address. \\
        & \textbf{Human:} & Thanks. I will also need a taxi from \textbf{the hotel} to \textbf{the restaurant}. Will you handle this?\\
        \cline{2-3}
        & \textbf{Gold DS:} & \begin{tabular}[c]{@{}l@{}}(hotel-area, west), (hotel-price\_range, cheap), (hotel-type, guest house),\\ (hotel-internet, yes), (hotel-name, warkworth house), (restaurant-area, centre),\\ (restaurant-food, italian), (restaurant-price\_range: cheap), (restaurant-name, \text{ask}),\\ (taxi-departure, warkworth house), (taxi-destination, \textbf{ask})\end{tabular}  \\
        & \textbf{CSFN-DST (no SG):} & \begin{tabular}[c]{@{}l@{}}(hotel-area, west), (hotel-price\_range, cheap), (hotel-type, guest house),\\ (hotel-internet, yes), (hotel-name, warkworth house), (restaurant-area, centre),\\ (restaurant-food, italian), (restaurant-price\_range: cheap), (restaurant-name, \text{ask}),\\ (taxi-departure, warkworth house), (taxi-destination, \textbf{warkworth house})\end{tabular} \\
        & \textbf{CSFN-DST:} & \begin{tabular}[c]{@{}l@{}}(hotel-area, west), (hotel-price\_range, cheap), (hotel-type, guest house),\\ (hotel-internet, yes), (hotel-name, warkworth house), (restaurant-area, centre),\\ (restaurant-food, italian), (restaurant-price\_range: cheap), (restaurant-name, \text{ask}),\\ (taxi-departure, warkworth house), (taxi-destination, \textbf{ask})\end{tabular} \\
        \hline
       \multirow{18}{*}{\textbf{C3}} & \textbf{Previous DS:} & \begin{tabular}[c]{@{}l@{}}(attraction-area, east), (attraction-name, funky fun house), (restaurant-area, east),\\ (restaurant-food, indian), (restaurant-price\_range, moderate),\\ (restaurant-name, \textbf{curry prince})\end{tabular}\\
       & \textbf{System:} & cb58jj is there postcode. Their address is 451 newmarket road fen ditton. \\
        & \textbf{Human:} & \begin{tabular}[c]{@{}l@{}}Great, thank you! Also, can you please book me a taxi between the restaurant and funky\\ fun house? I want to leave \textbf{the restaurant} by 01:30.\end{tabular}\\
        \cline{2-3}
        & \textbf{Gold DS:} & \begin{tabular}[c]{@{}l@{}}(attraction-area, east), (attraction-name, funky fun house), (restaurant-area, east),\\ (restaurant-food, indian), (restaurant-price\_range, moderate),\\ (restaurant-name, \text{curry prince}), (taxi-departure, \textbf{curry prince}),\\ (taxi-destination, funky fun house), (taxi-leave\_at, 01:30)\end{tabular}  \\
        & \textbf{CSFN-DST (no SG):} & \begin{tabular}[c]{@{}l@{}}(attraction-area, east), (attraction-name, funky fun house), (restaurant-area, east),\\ (restaurant-food, indian), (restaurant-price\_range, moderate),\\ (restaurant-name, \text{curry prince}), (taxi-departure, \textbf{curry garden}),\\ (taxi-destination, funky fun house), (taxi-leave\_at, 01:30)\end{tabular} \\
        & \textbf{CSFN-DST:} & \begin{tabular}[c]{@{}l@{}}(attraction-area, east), (attraction-name, funky fun house), (restaurant-area, east),\\ (restaurant-food, indian), (restaurant-price\_range, moderate),\\ (restaurant-name, \text{curry prince}), (taxi-departure, \textbf{curry prince}),\\ (taxi-destination, funky fun house), (taxi-leave\_at, 01:30)\end{tabular} \\
        \hline
        
       \multirow{13}{*}{\textbf{C4}} & \textbf{Previous DS:} & \begin{tabular}[c]{@{}l@{}}(hotel-name, \textbf{a and b guest house}), (hotel-book\_day, tuesday), (hotel-book\_people, 6),\\ (hotel-book\_stay, 4), (attraction-area, west), (attraction-type, museum)\end{tabular}\\
       & \textbf{System:} & \begin{tabular}[c]{@{}l@{}}\textbf{Cafe jello gallery} has a free entrance fee. The address is cafe jello gallery, 13 magdalene\\ street and the post code is cb30af. Can I help you with anything else?\end{tabular} \\
        & \textbf{Human:} & Yes please. I need a taxi to \textbf{commute}.\\
        \cline{2-3}
        & \textbf{Gold DS:} & \begin{tabular}[c]{@{}l@{}}(hotel-name, a and b guest house), (hotel-book\_day, tuesday), (hotel-book\_people, 6),\\ (hotel-book\_stay, 4), (attraction-area, west), (attraction-type, museum),\\ (taxi-destination, \textbf{cafe jello gallery}), (taxi-departure, \textbf{a and b guest house})\end{tabular}  \\
        & \textbf{CSFN-DST (no SG):} & \begin{tabular}[c]{@{}l@{}}(hotel-name, a and b guest house), (hotel-book\_day, tuesday), (hotel-book\_people, 6),\\ (hotel-book\_stay, 4), (attraction-area, west), (attraction-type, museum)\end{tabular} \\
        & \textbf{CSFN-DST:} & \begin{tabular}[c]{@{}l@{}}(hotel-name, a and b guest house), (hotel-book\_day, tuesday), (hotel-book\_people, 6),\\ (hotel-book\_stay, 4), (attraction-area, west), (attraction-type, museum),\\ (taxi-destination, \textbf{cafe jello gallery})\end{tabular} \\
        \hline
    \end{tabular}
    \caption{Four cases on the test set of MultiWOZ 2.1. DS means Dialogue State, and SG means Schema Graph.}
    \label{tab:case_study}
\end{table*}

\end{document}